\documentclass[pdflatex,sn-mathphys-num,iicol]{sn-jnl}

\usepackage{graphicx}%
\usepackage{multirow}%
\usepackage{amsmath,amssymb,amsfonts}%
\usepackage{amsthm}%
\usepackage{mathrsfs}%
\usepackage[title]{appendix}%
\usepackage{xcolor}%
\usepackage{textcomp}%
\usepackage{manyfoot}%
\usepackage{booktabs}%
\usepackage{algorithm}%
\usepackage{algorithmicx}%
\usepackage{algpseudocode}%
\usepackage{listings}%
\usepackage[caption=false, font=normalsize,labelfont=sf,textfont=sf]{subfig}
\usepackage{booktabs} 
\usepackage{pifont}
\newcommand{\cmark}{\ding{51}} 

\theoremstyle{thmstyleone}%
\theoremstyle{thmstyletwo}%

\theoremstyle{thmstylethree}%

\begin{document}

\title[Article Title]{An Interactive Conversational 3D Virtual Human}

\author[]{\fnm{Richard} \sur{Shaw}}\email{richard.shaw1@huawei.com}
\equalcont{These authors contributed equally to this work.}

\author[]{\fnm{Youngkyoon} \sur{Jang}}\email{youngkyoonjang@huawei.com}
\equalcont{These authors contributed equally to this work.}

\author[]{\fnm{Athanasios} \sur{Papaioannou}}\email{athanasios.papaioannou@huawei.com}
\equalcont{These authors contributed equally to this work.}

\author[]{\fnm{Arthur} \sur{Moreau}}\email{arthur.moreau3@huawei.com}
\equalcont{These authors contributed equally to this work.}

\author[]{\fnm{Helisa} \sur{Dhamo}}\email{helisa.dhamo@huawei.com}
\equalcont{These authors contributed equally to this work.}

\author[]{\fnm{Zhensong} \sur{Zhang}}\email{zhangzhensong@huawei.com}
\equalcont{These authors contributed equally to this work.}

\author[]{\fnm{Eduardo} \sur{P\'erez-Pellitero}}\email{e.perez.pellitero@huawei.com}
\equalcont{These authors contributed equally to this work.}

\affil[]{\orgdiv{Noah's Ark Lab}, \orgname{Huawei}, \city{London}, \country{UK}}


\abstract{This work presents \textit{Interactive Conversational 3D Virtual Human} (ICo3D), a method for generating an interactive, conversational, and photorealistic 3D human avatar. Based on multi-view captures of a subject, we create an animatable 3D face model and a dynamic 3D body model, both rendered by splatting Gaussian primitives. Once merged together, they represent a lifelike virtual human avatar suitable for real-time user interactions. We equip our avatar with an LLM for conversational ability. During conversation, the audio speech of the avatar is used as a driving signal to animate the face model, enabling precise synchronization. We describe improvements to our dynamic Gaussian models that enhance photorealism: SWinGS++ for body reconstruction and HeadGaS++ for face reconstruction, and provide as well a solution to merge the separate face and body models without artifacts. We also present a demo of the complete system, showcasing several use cases of real-time conversation with the 3D avatar. Our approach offers a fully integrated virtual avatar experience, supporting both oral and written form interactions in immersive environments. ICo3D is applicable to a wide range of fields, including gaming, virtual assistance, and personalized education, among others. \textbf{Project page:} \url{https://ico3d.github.io/}
}


\maketitle

\footnotetext{This preprint has not undergone peer review or any post-submission improvements or corrections. The Version of Record of this article is published in \emph{International Journal of Computer Vision} and is available online at \url{https://doi.org/10.1007/s11263-025-02725-8}.}

\section{Introduction}\label{sec1}

The development of interactive 3D virtual humans has gained significant traction in recent years, driven by advancements in artificial intelligence, computer vision, and real-time rendering technologies. These digital avatars are becoming integral to a wide range of applications, including gaming, entertainment, education, healthcare, and customer service. By enabling realistic, dynamic interactions between users and virtual characters, 3D virtual human avatars have the potential to redefine the way users experience virtual or augmented environments. 
Capturing, reconstructing, and reanimating an interactive virtual avatar that moves and behaves like a real human is a complex problem composed of a multiplicity of interconnected tasks, such as face reconstruction, speech and expression reanimation, body reconstruction and pose control, natural language understanding and generation. Prior art has generally 
focused on solving each of these problems individually, and often independently, in part as capturing multi-view sequences of body and head requires different hardware arrangements in terms of space, cameras and illumination.

\begin{figure}[]
\centering
\includegraphics[width=0.45\textwidth]{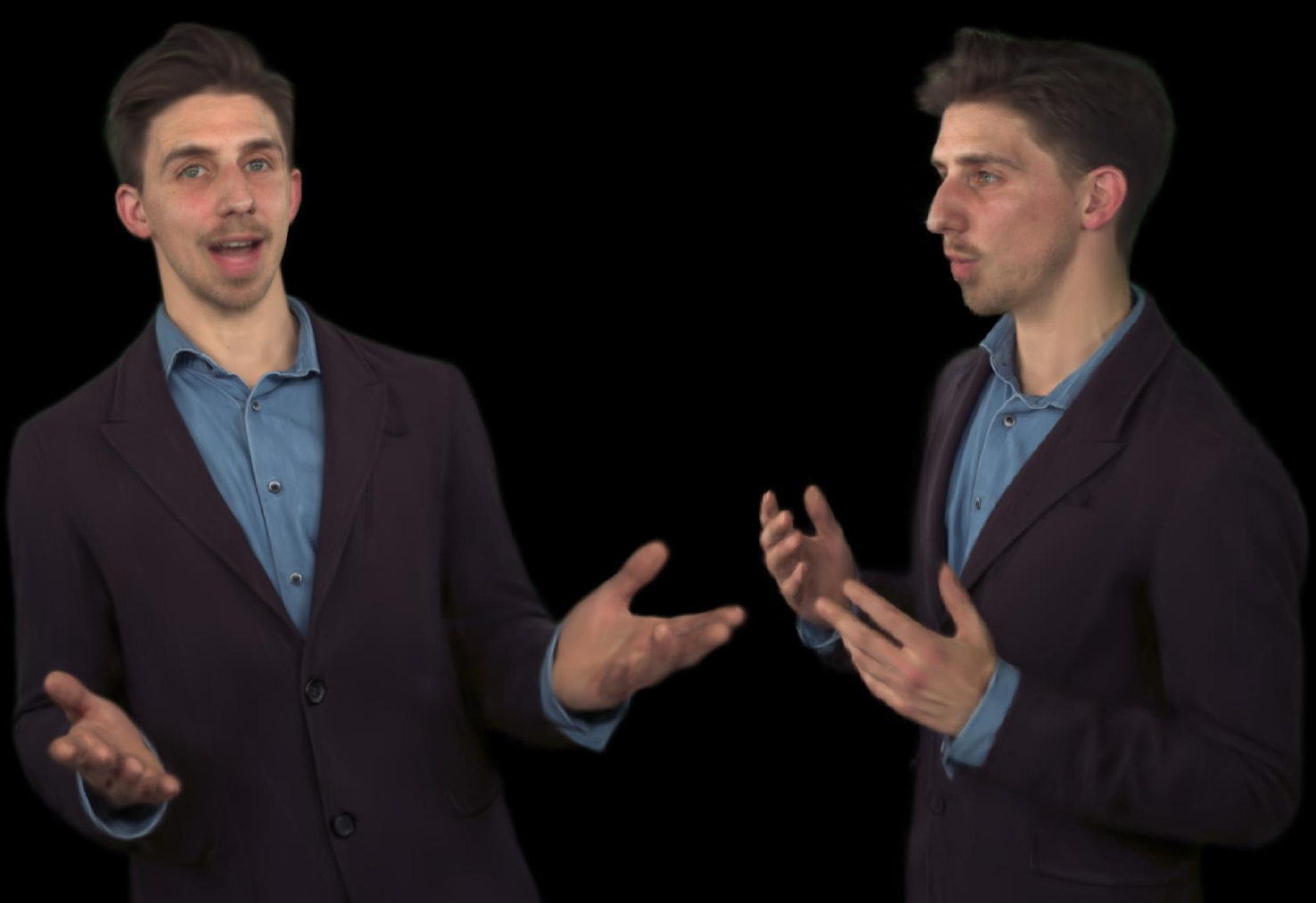}
\caption{Our method generates a photorealistic 3D full-body virtual avatar from posed multi-view images, with dynamic facial expressions driven by input audio and corresponding procedural body animation. A LLM enables the user to converse with the avatar in real-time. We refer the reader to our suppl.\,video to see our method in action.}
\label{fig:teaser1}
\end{figure}

Arguably, the face is the most attention-attracting element of a 3D virtual human avatar. A number of recent works ~\cite{Qian2023GaussianAvatarsPH, xiang2024flashavatar, Zielonka2022InstantVH, xu2023gaussianheadavatar} have obtained impressive results in photorealistic reconstruction of heads and faces, with increasingly faster rendering times using both monocular or multi-view input videos. Often, they retain a control interface for new expressions via parametric 3D Morphable Models (3DMM), such as FLAME~\cite{Li2017LearningAM} and ARKit~\cite{ARKit}. Despite their success, these works are generally disconnected from the full body, sometimes even lacking neck articulation. Additionally, the expression reanimation often relies on pre-recorded face tracking or manually edited 3D blendshapes, making them not easily generative.

In this work, we build upon our previous work on head~\cite{dhamo2023headgas} and dynamic \cite{shaw2024swings} 3D body reconstruction. We introduce technical improvements over both methods, and build a full-bodied conversational avatar that runs in real time. 

Our contributions are: \textbf{(1)} extend our head model HeadGaS to enable audio-driven head reanimation (Sec.\,\ref{sec:head_model}), and \textbf{(2)} propose a novel intra-window spatio-temporal encoder that enhances temporal consistency and accuracy of our dynamic body model SWinGS (Sec.\,\ref{sec:body_eval}). \textbf{(3)} We then integrate both head and body reconstruction to produce a complete avatar (Sec.\,\ref{sec:head_body_integ}), and \textbf{(4)} introduce a Large Language Model (LLM) that provides interactive conversational capabilities (Sec.\,\ref{sec:llm_setup}). Given the text response from the LLM, a Text-to-Speech (TTS) model generates speech audio that drives our head model, enabling realistic synchronized facial animations. Thanks to the inherent efficiency of 3D Gaussian Splatting at inference time, we can render the avatar in our real-time viewer, allowing users to chat with it while wearing a VR headset or watching a screen.

The remainder of this paper is organized as follows: In Sec.~\ref{sec:related_work}, we review related
work in dynamic 3D human reconstruction, head reconstruction and animation, and Natural Language Processing (NLP). In Sec.~\ref{sec:method}, we describe the building blocks of our conversational virtual human, followed by Sec.~\ref{sec:implementation_details} with further implementation details, specially relating to our real-time viewer. We present quantitative and qualitative results of our algorithms and system in Sec.~\ref{sec:experiments}. We conclude the paper in Sec.~\ref{sec:conclusions}.

\section{Related Work}\label{sec:related_work}

\subsection{Neural Rendering of Dynamic Human Bodies}

Recent advancements in neural rendering have significantly improved the fidelity, photorealism, and efficiency of 3D human avatars compared to commonly used textured meshes. Pioneering works in human body modelling use statistical mesh templates such as SMPL~\cite{SMPL:2015,SMPL-X:2019}, which are convenient to use but lack flexibility to represent the exact body shape and appearance of a person, and thus are often used as building blocks of more advanced methods. 

Since then, another important breakthrough has been the seminal work of Neural Radiance Fields (NeRF)~\cite{Mildenhall2020NeRF} that enabled photorealistic rendering of static scenes via volume rendering. The computationally intensive first version has quickly been improved enabling real-time rendering by storing spatially distributed features~\cite{Mller2022InstantNG,Yu2021PlenOctreesFR}. NeRF has also been extended to dynamic scenes by modeling the time dimension~\cite{li2020neural, pumarola2020d, park2021nerfies}. These dynamic radiance fields can be used to train and render volumetric representation of humans in motion~\cite{peng2023representing,isik2023humanrf}. Alternatively, one can also use implicit surface representations~\cite{Wang2021NeuSLN, Takikawa2021NeuralGL} to render a 3D human~\cite{Saito2020PIFuHDMP, Jiang_2022_CVPR}.

Recently, 3D Gaussian Splatting (3DGS)~\cite{kerbl3Dgaussians} made significant strides by modeling scenes with 3D Gaussian primitives, which, when combined with tile-based differentiable rasterization, achieves very fast rendering while maintaining high-quality reconstruction. Luiten et al.~\cite{luiten2024} extended this to dynamic scenes with tracked 3D Gaussian trajectories. Recent methods extend 3DGS to general dynamic scenes including~\cite{yang2023gs4d, yang2023deformable3dgs, huang2023sc, lin2023gaussian,  li2023spacetime, sun20243dgstream} among others. Other works have focused specifically on adapting 3DGS to dynamic human reconstruction, including~\cite{moreau2024human, li2024animatable, kocabas2024hugs, hu2024gaussianavatar, pang2024ash, qian20233dgsavatar, Wu_2024_CVPR, wen2024gomavatar,jung2023deformable}. 

Regardless of the representation (e.g.\,mesh, NeRF, SDF, 3D Gaussians), one important design choice (explored in Sec.~\ref{sec:hugs_comparison}) is the driving signal for the body motion: learning time-dependent motion leads to 4D models that can replay the training videos~\cite{peng2023representing,isik2023humanrf,shaw2024swings}, while using 3D body pose as input enables the creation of animatable body models~\cite{bagautdinov2021driving, moreau2024human, li2024animatable, kocabas2024hugs, pang2024ash} deformed with skinning algorithms~\cite{merry2006animation,kavan2007skinning}.

Our method specifically extends upon our prior work SWinGS~\cite{shaw2024swings} for multi-view 4D reconstruction. By incorporating recent advancements, we enable real-time and higher-fidelity full-body 3D reconstructions, providing a comprehensive solution for our interactive virtual human avatar.

\subsection{Head Reconstruction and Facial Animation}
Head reconstruction from a set of image observations has been a very active field in recent years. Approaches include models that generalize across subjects~\cite{Hong2021HeadNeRFAR, Mihajlovic2022KeypointNeRFGI, Wang2022MoRFMR} or rely on multi-view head captures~\cite{YoungkyoonJang2023VSCHH2A, Kirschstein2023NeRSembleMR, Lombardi2018DeepAM, Lombardi2021MixtureOV}, which can have a static or dynamic form. The task of generating animatable facial models from images has progressed significantly, with recent works enabling high-fidelity facial reenactments and dynamic expressions. Deep Video Portraits~\cite{Kim2018DeepVP} and Face2Face~\cite{Thies2016Face2FaceRF} enable real-time facial reenactment with high realism but rely on constrained source data, such as a frontal single-view image and limited expressions. Voice2Face~\cite{Aylagas2022Voice2FaceAF} employs audio signals for facial animation, but the expressions and conversational integration are limited.

Recent 3D Gaussian Splatting-based approaches to facial animation modelling include~\cite{Qian2023GaussianAvatarsPH, xiang2024flashavatar, saito2024rgca, xu2023gaussianheadavatar}. Our method extends our prior work HeadGaS~\cite{dhamo2023headgas} by integrating audio-driven facial animation with generative, LLM-based conversational capabilities. This enables our avatars to reason and generate language, and to exhibit contextually appropriate facial expressions and respond to user inputs with realistic movements in real time.

\subsection{Natural Language Processing and Conversational Agents}
Language is central to the way humans acquire, organize and communicate knowledge. Language modelling has been a long-standing field that aims to provide computational models that can understand and generate diverse human language \cite{devlin-bert, gao2004}. Language Models (LMs), such as the ubiquitous \textit{n}-gram model \cite{brown1992}, generally predict the likelihood of word sequences based on the surrounding context. Recently, Large Language Models (LLMs) have shown impressive performance in text understanding and generation. The term LLM refers to a family of models with a very large parameter count, which have been trained on an immense corpus of data, i.e.\,at web-scale. Notable examples of LLMs such as GPT-3 \cite{floridi_cgpt3_2020}, InstructGPT \cite{ouyang_2024}, and GPT-4, \cite{gpt4_2024} rely on the \textit{Transformer} architecture~\cite{vaswani_2017}, which, thanks to the self-attention mechanism, can flexibly capture context and long-range dependencies. In part due to their scale advantage, LLMs present strong generalization and unprecedented naturalness across different tasks and domains, specially those related to conversational capabilities, e.g. natural language understanding, reasoning, and generation. We refer the reader to Yupen et al.~\cite{yupeng_llm_survey_2024} for a more exhaustive literature review on LLMs. 

In this work, we leverage publicly available LLMs to provide advanced conversational capabilities to our 3D human reconstruction, which effectively provides a 3D embodiment of the LLM. Qwen~\cite{qwen} provides 
an open-source LLM series consisting of various versions ($1.5$, $2$, $2.5$) and number of parameters (ranging from $0.5$B to $110$B) \cite{zhao2023survey}. In our pipeline, we deployed Qwen2 which supports multiple  languages, including  English and Chinese, and whose model weights are available on Hugging Face and ModelScope.

\section{Method}\label{sec:method}
\subsection{System Overview}

\begin{figure*}[]
\centering
\includegraphics[width=0.99\textwidth,trim={0 25 0 5},clip]{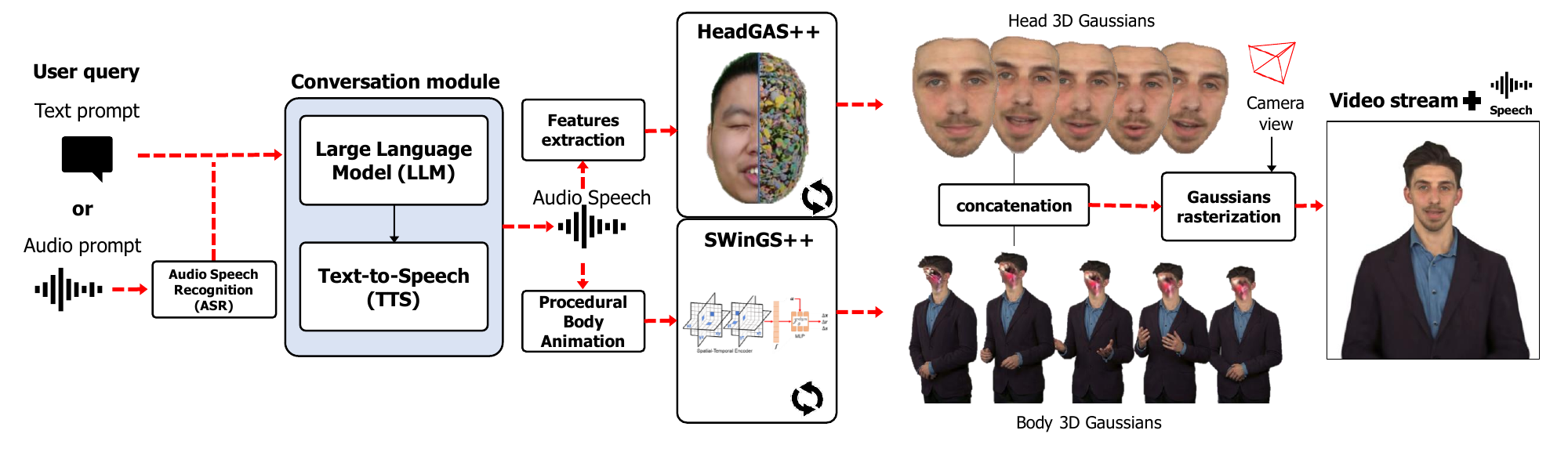}
\caption{\textbf{Overview of our conversational human pipeline ICo3D.} Users interact with the avatar via text or audio queries, which are processed by a LLM to produce a textual answer, then converted to audio. The audio speech is the driving signal for the audio-driven head model \textit{HeadGaS++} (Fig.~\ref{fig:headgas}). It also serves to determine the body dynamics through procedural body animation. 3D Gaussians are generated at each timestep by both the head and the body model \textit{SWinGS++} (Fig.~\ref{fig:swings}), which are then integrated and rendered from novel views, producing a free viewpoint video stream synchronized with the audio speech.}
\label{fig:overview}
\end{figure*}

Our complete system for building a full-body animatable and interactive 3D virtual avatar integrates separate reconstruction methods for head and body modeling, along with LLM integration for driving real-time facial animation. We divide the reconstruction process into three main components: the audio-driven animatable head model (Sec.~\ref{sec:head_model}), the dynamic body model (Sec.~\ref{sec:body_model}), and head-body integration (Sec.~\ref{sec:head_body_integ}). The head model is first trained separately to obtain high-quality renderings and to learn person-specific facial expressions from the data. It is then fused with the body model during the integration process to generate the complete avatar. Each reconstruction method is built upon the 3D Gaussian Splatting framework.

\subsection{3D Gaussian Splatting}

The original 3D Gaussian Splatting (3DGS) method~\cite{kerbl3Dgaussians} models a static scene using a set of 3D Gaussians. Given a set of images and associated camera poses, 3DGS initializes its representation with a sparse point cloud, typically generated using the COLMAP structure-from-motion pipeline~\cite{Schnberger2016StructurefromMotionR}. Each 3D Gaussian is represented by its center $\boldsymbol{\mu} \in \mathbb{R}^{3}$, covariance matrix ${\boldsymbol{\Sigma}} \in \mathbb{R}^{3 \times 3}$, opacity $o \in \mathbb{R}$, and color $c \in \mathbb{R}^{3(k+1)^2}$, where $k$ denotes the degree of spherical harmonics used to model the color, capturing view-dependent appearance. The Gaussians are placed in world coordinates and evaluated at a point $\boldsymbol{x}$ using the following equation:

\begin{equation}
    G(\boldsymbol{x}) = e^{-\frac{1}{2} \boldsymbol{x}^T {\boldsymbol{\Sigma}}^{-1} \boldsymbol{x}} .
\end{equation}

The covariance matrix $\mathbf{\Sigma}$ is computed as $\boldsymbol{\Sigma} = \mathbf{R} \mathbf{S} \mathbf{S}^T \mathbf{R}^T$, where $\mathbf{S}$ and $\mathbf{R}$ are the scaling and rotation matrices, respectively. To represent the scene, 3DGS optimizes the parameters of the 3D Gaussians using differentiable rasterization and Gaussian Adaptive Density Control (ADC). The loss function used for training is defined as:

\begin{equation}
    \mathcal{L} = (1-\lambda)\mathcal{L}_1(I_r, I_{gt}) + \lambda \mathcal{L}_{\mathrm{SSIM}}(I_r,I_{gt}),
\end{equation}

\noindent where $I_r$ and $I_{gt}$ denote the rendered image and ground-truth image, respectively. This framework is highly effective for static scenes but lacks the ability to handle dynamics, such as the nuanced facial expressions and body movements of our avatar, which we address in the subsequent sections.

\subsection{HeadGaS++: Audio-driven 3D Head Model}
\label{sec:head_model}

\begin{figure*}[t]
\centering
\includegraphics[width=0.99\textwidth]{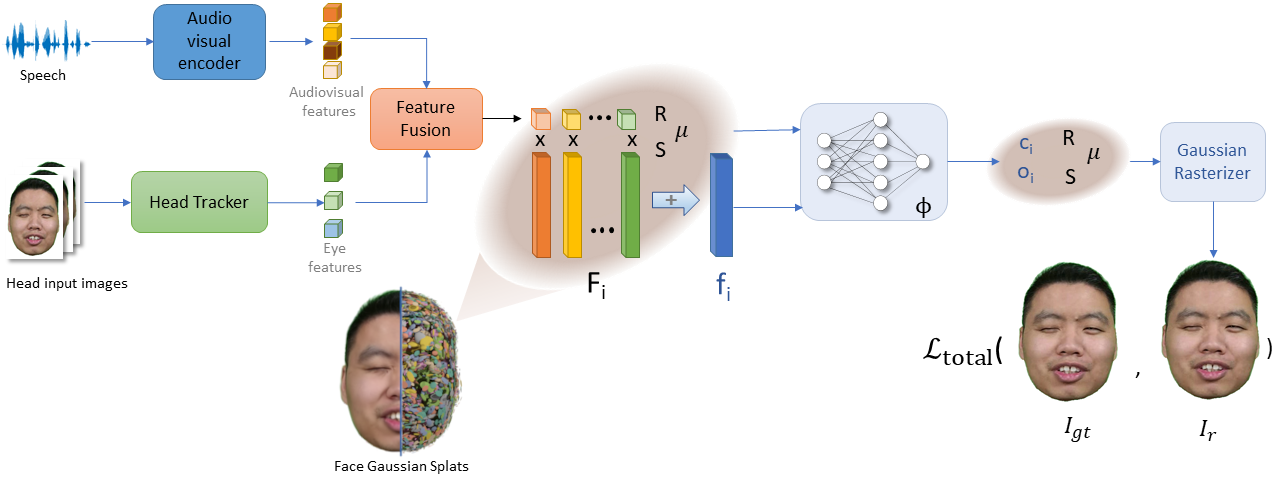}
\caption{\textbf{An overview of our method for 3D animatable head reconstruction.} \textit{HeadGaS++}, is an extension of our work HeadGaS~\cite{dhamo2023headgas}, which uses features extracted from input audio speech to drive the facial expressions.}
\label{fig:headgas}
\end{figure*}

In this section, we describe our method HeadGaS++, our approach to generating high-fidelity, audio-driven 3D head models from multi-view video sequences and speech inputs. The goal is to synthesize novel views and expressions of the avatar's face given a novel audio feature vector and camera pose as input. An overview of our method is shown in Fig.~\ref{fig:headgas}. The method is an extension of our prior work on animatable head modelling HeadGaS~\cite{dhamo2023headgas}, but with enhanced audio-driven capabilities and higher fidelity reconstructions (see Table~\ref{tab:head_quantitative}).

As in HeadGaS, HeadGaS++ extends the static 3DGS framework to create animatable avatars by introducing audio-dependent Gaussian properties. We enable each Gaussian to change its color $c_i$ and opacity $o_i$ based on the input audio-visual features. Rather than moving the Gaussian positions directly, this framework results in 3D Gaussians with dynamic appearances. These Gaussians can occasionally appear and vanish depending on the current audio-visual features, and additionally allow color changes to simulate non-rigid appearance effects.

To this end, we augment every Gaussian primitive with a learned latent feature basis $\mathbf{F}\in\mathbb{R}^{B\times f} $, as shown in Fig.~\ref{fig:headgas} (Audiovisual Feature Basis), where $B$ is the number of blendshape expressions. This feature basis is blended with an audio-visual feature vector to enable dynamic expression control. During training, the latent basis $\mathbf{F}$ is optimized together with the other parameters of the 3D Gaussians. At each iteration, we blend the respective audio-visual weights $\mathbf{e}_{i} \in \mathbb{R}^B$ corresponding to a particular expression at the current frame index $i$, with the feature basis $\mathbf{F}$ into a 1D feature vector $\mathbf{f}_i\in\mathbb{R}^f$ as:
\begin{equation}
\mathbf{f}_i = \mathbf{F}^{T} \mathbf{e}_i + \mathbf{f}_0 ,
\end{equation}

\noindent where $\mathbf{f}_0$ is a bias term. Then, feature $\mathbf{f}_i$ is fed into a small MLP
$\phi(\cdot)$, to compute the Gaussian's color $c_i$ and opacity $o_i$:
\begin{equation}
    c_i, o_i = \phi(\mathbf{f}_i, \gamma(\boldsymbol{\mu})) ,
\end{equation}

\noindent where $\gamma$ denotes the sinusoidal positional encoding function, and $\boldsymbol{\mu}$ denotes the Gaussian mean position.  

We use a compact MLP composed of two linear layers with LeakyReLU activation; the hidden layer has 64 channels, and the last layer consists of two branches for color and opacity prediction. The model is optimized by comparing the rendered image $I_r$ with the ground truth image $I_{gt}$, using the following loss function:
\begin{equation}
    L = \lambda_{1} L_{1}(I_r, I_{gt}) + \lambda_{s} L_{s}(I_r, I_{gt}) + \lambda_{p} L_{p}(I_r, I_{gt}),
\end{equation}

\noindent where, $\lambda_{1}, \lambda_{s}$, and $\lambda_{p}$ are weighting factors, and $L_{s}$ and $L_{p}$ are SSIM and perceptual losses~\cite{Johnson2016PerceptualLF}, respectively.

\subsection{SWinGS++: Dynamic 3D Body Reconstruction}
\label{sec:body_model}

\begin{figure*}[]
\centering
\includegraphics[width=0.99\textwidth]{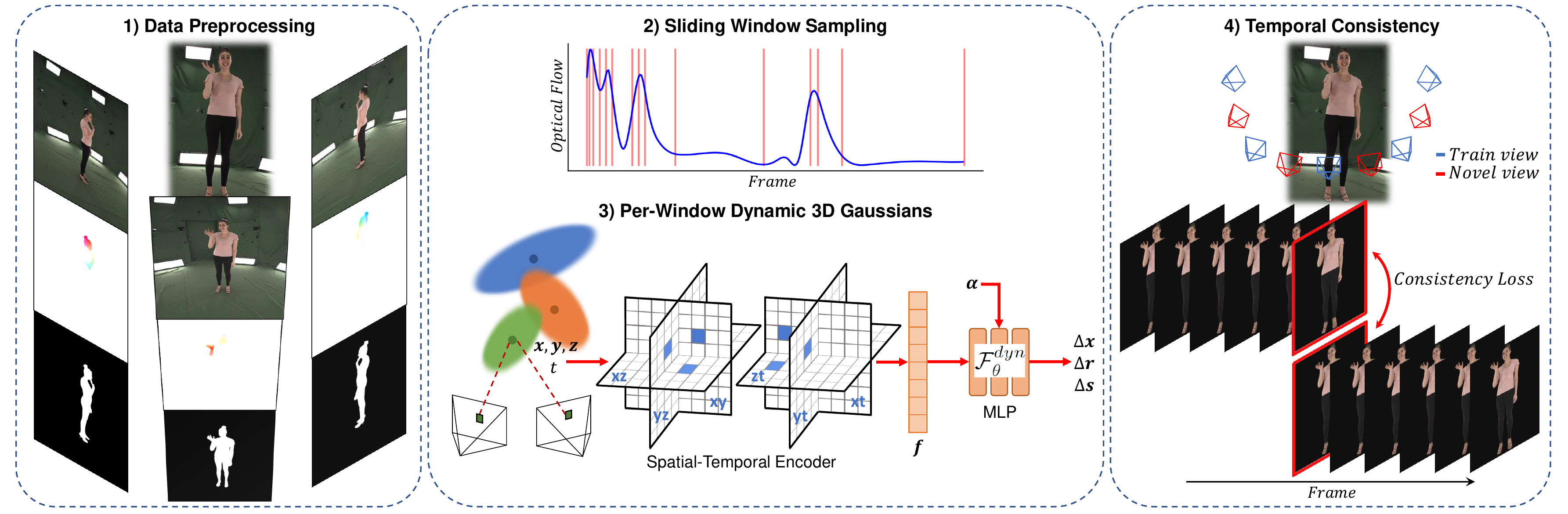}
\caption{\textbf{An overview our dynamic 3DGS method for body reconstruction.}. \textit{SWinGS++} is an extension of our work SWinGS~\cite{shaw2024swings}, using temporally-local dynamic MLPs on a sliding window basis. The method is extended using a spatial-temporal encoder to help reconstruct larger or faster human movements.}
\label{fig:swings}
\end{figure*}

Once the animatable head model is trained, our focus shifts to 4D body reconstruction to complete the avatar. To achieve the level of fidelity required to create a convincing and realistic virtual human, we extend our prior work SWinGS~\cite{shaw2024swings}. SWinGS enables high-quality rendering of general dynamic scenes given multi-view, calibrated and time-synchronized videos as input. SWinGS++ extends the method by shifting the focus to dynamic humans with the introduction of a spatial-temporal encoder to provide better human motion estimation. An overview of our extended method is shown in Fig.~\ref{fig:swings}.

The body reconstruction pipeline comprises three main components: 1) We divide an input sequence into overlapping variable-length sliding windows, based on the human's motion. This allows us to handle long sequences while preserving render quality. 2) We train independent 4DGS models for each sliding window, where per-window tunable MLPs~\cite{Maggioni2023TunableCW} learn deformations from temporally-local canonical Gaussians to each frame. This allows us to capture significant geometric changes as well as the emergence of new objects.

3) A final temporal fine-tuning stage enforces temporal consistency throughout the sequence.

We discuss each of these steps in the following sections.

\subsubsection{Sliding Window Sampling} 
Representing the dynamic human with a single 4DGS model is impractical, in terms of computational load and performance degradation, especially with fast human motion and long sequences. To address this, following~\cite{shaw2024swings}, we introduce a sliding-window strategy; partitioning the sequence into smaller overlapping windows. Each window is an independent 4DGS model, allowing all Gaussian parameters--positions, rotations, scaling, colors, and opacities--to change between windows. Our adaptive sampling method dynamically adjusts the window size based on the human's motion, balancing training efficiency and performance. Given a sequence of length $N_f$ frames, we split it into a set of sliding windows of lengths $\{N_{w}\}$, depending on the motion: high-motion regions are sampled more frequently (shorter windows), low-motion regions are sampled less frequently (longer windows). The motion is estimated using 2D optical flow from a pre-trained RAFT model~\cite{Teed2020RAFTRA}. For each camera view $c \in C$, we accumulate the optical flow magnitude $\hat{v}_i$ across frames:

\begin{equation}
\hat{v}_i = \frac{1}{C}
    \sum^{C}_c 
    \sum^{N_f-1}_i
     || \boldsymbol{v} (I^c_i, I^c_{i+1} ) ||^2_2,
\end{equation}

\noindent and a new window is spawned at frame $i$ when the accumulated flow exceeds a predefined threshold $\hat{v}_i > v_{thresh}$. 

\subsubsection{Dynamic 3D Body Gaussians}
To extend 3DGS for dynamic scenes, \cite{shaw2024swings} introduced temporally-local dynamic MLPs $\phi(\cdot)$ for each sliding window, where each MLP learns the deformation mapping from a per-window canonical 3D Gaussian representation to the set of 3D Gaussians for each frame $i$ within the window. We use a compact MLP which takes as input the normalized frame time $t \in [0,1]$ and 3D Gaussian positions $\boldsymbol{\mu}$ (normalized by the scene's mean and standard deviation), and outputs displacements to their positions $\Delta \boldsymbol{\mu}$, rotations $\Delta \boldsymbol{r}$ and scaling $\Delta \boldsymbol{s}$ at each time step:

\begin{equation}
    \Delta \boldsymbol{\mu}(t), \Delta \boldsymbol{r}(t), \Delta \boldsymbol{s}(t)
    = \phi(\gamma({\boldsymbol{\mu}}), \gamma(t)).
\end{equation}

However, we found this model insufficient to capture human motion when the inter-frame displacement is large. Therefore, SWinGS++ extends the method by introducing per-window spatial-temporal encoders, similar to \cite{FridovichKeil2023KPlanesER, Wu_2024_CVPR}, as shown in Fig.~\ref{fig:swings}. The input 3D position $\boldsymbol{\mu}$ and time $t$ are passed through the spatial-temporal encoder $\mathcal{H}$ producing voxel features $\mathbf{f}_v = \mathcal{H}(\boldsymbol{\mu},t)$, which are then passed to the MLP:

\begin{equation}
    \label{eq:swings_mlp}
    \Delta \boldsymbol{\mu}(t), \Delta \boldsymbol{r}(t), \Delta \boldsymbol{s}(t)
    = \phi ( \mathbf{f}_v ) ,
\end{equation}

\noindent resulting in better motion estimation and reconstruction, as shown in Table~\ref{tab:body_nvs} and our qualitative results in Fig.~\ref{fig:novel_view}.

The MLP uses the same architecture as \cite{shaw2024swings}, i.e. a \textit{tunable MLP}~\cite{Maggioni2023TunableCW} consisting of $M$ sets of weights with blending parameters $\{ \boldsymbol{\alpha} \in \mathbb{R}^{M \times N_g} \}$. These sets of weights enable the MLP to learn $M$ \textit{motion modes}, smoothly weighted by blending parameters $\boldsymbol{\alpha}$ for each input Gaussian. The output of a single layer of the dynamic MLP can be expressed as:

 \begin{equation}
    \boldsymbol{y} = \psi \left( \sum^M_{m=1} \left( \alpha_{m} \boldsymbol{w}_m^T \boldsymbol{x} + \alpha_{m} \boldsymbol{b}_m \right) \right),
\end{equation}

\noindent where $\boldsymbol{w}$ and $\boldsymbol{b}$ are the weights and bias, and $\psi$ is an activation function. 
Thus, the MLP from Eq.~(\ref{eq:swings_mlp}) becomes $\phi_{dyn}$, a function of spatial-temporal features $\mathbf{f}_v$ and blending parameters $\boldsymbol{\alpha}$:

\begin{equation}
    \Delta \boldsymbol{\mu}(t), \Delta \boldsymbol{r}(t), \Delta \boldsymbol{s}(t)
    = \phi_{dyn} (\mathbf{f}_v, \boldsymbol{\alpha}).
\end{equation}

We optimize the model using the following loss function:
\begin{equation}
    L = \lambda_{1} L_{1}(I_r, I_{gt}) + \lambda_{s} L_{s}(I_r, I_{gt}) + \lambda_{m} L_{m}(M_r, M_{gt}),
\end{equation}

\noindent where $L_{m}$ is a mask loss which is simply the $L_1$ loss on the rendered $M_r$ and ground truth $M_{gt}$ human masks.

\subsubsection{Temporal Consistency Fine-tuning}
After training each sliding window independently, optimization discrepancies lead to temporal artifacts on the body, such as flickering, mainly apparent in novel views. To mitigate this, we employ temporal fine-tuning to enforce inter-window consistency. We fine-tune each sliding window model, starting from the first and proceeding sequentially through the sequence.
We randomly sample novel views $\mathbf{P}_{novel}$ by interpolating the training camera poses in $\mathbf{SE(3)}$:

\begin{equation}
    \mathbf{P}_{novel} = \exp_{\mathrm{M}} \sum^{C}_c \beta_c \log_{\mathrm{M}} ( \mathbf{P}_c ) ,
\end{equation}

\noindent where $\exp_\mathrm{M}$ and $\log_\mathrm{M}$ are the matrix exponential and logarithm respectively, and $\beta_c \in [0,1]$ is a uniformly sampled weighting.
We apply consistency loss on novel view renders, which is simply the $L1$ loss on the overlapping frames of the current $w$ and previous $w-1$ windows:

\begin{equation}
    L_{consistency} = | I^w_{r,t=0} - I^{w-1}_{r,t=N_w-1} |_1 .
\end{equation}

During fine-tuning, we freeze the previous window's model and the MLPs, allowing only the canonical Gaussians to optimize. We use an alternating training strategy which optimizes for temporal consistency on overlapping frames for $75\%$ of the time, while using the remaining $25\%$ for standard training.

\subsection{Head-Body Model Integration}
\label{sec:head_body_integ}
Once the head model is trained (Sec.~\ref{sec:head_model}), the next step is to integrate the head with the body to produce a full-bodied avatar that can be rendered in real-time. The integration process happens jointly with the optimization of the body (Sec.~\ref{sec:body_model}) as this enables smooth unification of the two models.
Our integration method includes the following steps: 1) 3D head-body pose alignment (see Sec.~\ref{sec:cross_setup} on cross-setup alignment), 2) Gaussian merging, 3) Gaussian pruning and blending, and lastly 4) face color optimization (if necessary). In this section, we describe each of these processes in detail. 

\begin{figure*}
\centering
\small
\includegraphics[width=1.0\textwidth]{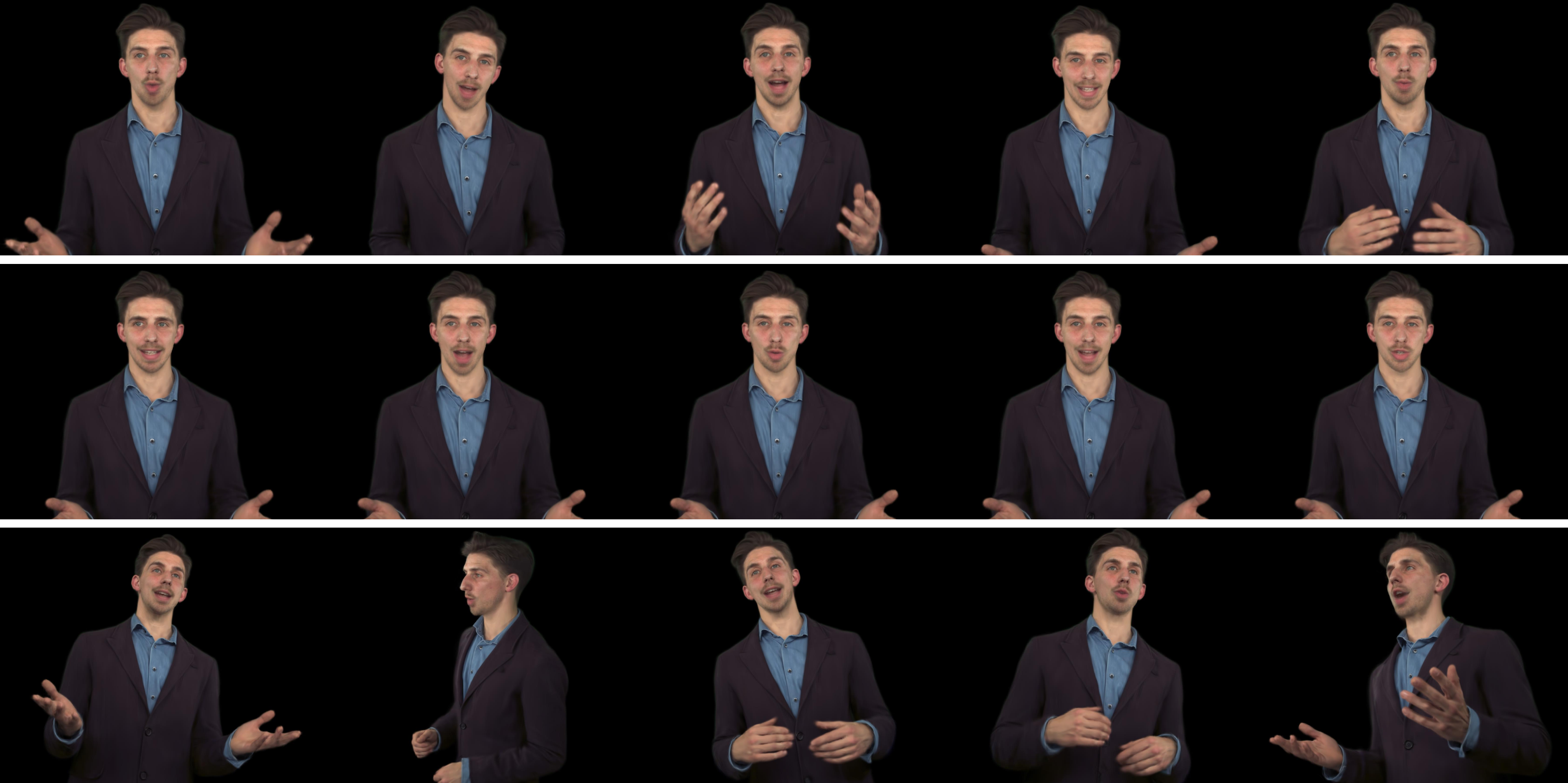}
\caption{\textbf{Qualitative results showing the interactive avatar's different modes of variation.} Top row: the avatar's facial animation is driven by novel expression parameters generated from input audio speech, while the body undergoes corresponding procedural animation. Middle row: a static human pose is held while the face is animated with novel expressions. Bottom row: the user has full free-viewpoint navigation of the fully animated 3D avatar in real time.}
\label{fig:qual_results}
\end{figure*}

First, we align the head model with the body model for each frame in the sequence. 
If the body and the face are visible 
in the same capture (as in Fig.~\ref{fig:qual_results}), the head and body models can be simply aligned using the tracked head poses, as the head and body segmentations are extracted from the same input images and both models use the same camera poses. In this case, the head Gaussians are transformed by the corresponding head pose $\mathbf{H}_i \in \mathbb{R}^{4 \times 4}$ at each frame $i$. In the case of head and body data from different sources, i.e. different capture setups (Fig.~\ref{fig:datasets}), the head must be rigidly aligned to the body with additional transformation matrices. This is explained in Sec.~\ref{sec:cross_setup}.

Secondly, to render the head and body models together, we merge their respective Gaussian representations into a unified representation by concatenating each of the Gaussian parameters together, before being fed into the rasterizer. This includes their positions, rotations, scales, colors and opacities. Thus, the complete avatar can be rendered efficiently with a single pass of the Gaussian Splatting rasterizer.

Thirdly, simply merging the Gaussians of each model together as stated results in significant artifacts. The reason is that Gaussians on the body model can penetrate the face model. Additionally, 2D head segmentation errors (or multi-view mask inconsistencies) can lead to artifacts at the face boundary, creating a noticeable \textit{transition} between face and body. To address these two issues, we apply Gaussian blending and pruning strategies. To prevent body Gaussians penetrating the face, we prune Gaussians from the body, every $n=100$ iterations, that lie within a spherical region positioned at the centroid of the face. We also prune Gaussians around the jaw to prevent ``double-chin'' artifacts appearing, e.g. when the face model opens the mouth but on the body it is closed. To address face border issues, at the start of training we inject randomly sampled Gaussians positioned around the border (extracted from the FLAME mesh) to help blend the edges.

The final integration stage is color or appearance optimization, i.e. we optimize for the colors of the face Gaussians to match the body. For captures with visible head and body, this step is not necessary as the head and body models are trained using the same input images and any differences in color due to optimization discrepancies are negligible. In the case of data from different sources with different lighting conditions, the appearance of the head must be adjusted to match the body. We do this by unfreezing the final layer of the HeadGaS++ model, i.e.\,the parameters that control the color of the Gaussians, and allow these to optimize concurrently with training of the body.

\subsection{Cross-setup Head Alignment}
\label{sec:cross_setup}

When training with the RenderMe-360~\cite{2023renderme360} and DNA-Rendering~\cite{cheng2023dna} datasets for head and body  reconstruction respectively, we face the challenge of having two separate captures of our avatar, acquired at different times with distinct capture setups (Fig.~\ref{fig:datasets}). Each setup has different world origins, camera poses and intrinsics, different motion sequences with changing head poses, facial expressions, and different lighting conditions, and other varying factors (e.g. appearance and clothing) that make integrating head and body challenging. 

For the task of cross-setup head alignment, the aim is to combine the 4DGS models so that the head can be rendered in the reference system of the body. In other words, transform the HeadGaS++ model trained with capture setup 1's head and camera poses (RenderMe-360), so it can be observed in capture 2's reference system (DNA-Rendering) and aligned to the per-frame head poses of capture 2. 
Generally, this would not have a closed-form solution, given that the captures are entirely independent. However, we can exploit the fact that the FLAME fittings from both sequences puts both heads (from capture setups 1 and 2) into a unified canonical reference system.

First, we can project a 3D point $\boldsymbol{x}$ from each of the 3D models into an image in their individual reference systems as:

\begin{align}
    \boldsymbol{p}_{w_1} &= \mathbf{K}_1 \mathbf{C}_{w_1}^{w2c} \mathbf{C}_{ref, w_1}^{-1} \mathbf{H}_{i,w_1} \boldsymbol{x}_{c} \\
    \boldsymbol{p}_{w_2} &= \mathbf{K}_2 \mathbf{C}_{w_2}^{w2c} \mathbf{C}_{ref, w_2}^{-1} \mathbf{H}_{i,w_2} \boldsymbol{x}_{c},
\end{align}

\noindent where $\boldsymbol{p}$ is a 2D image point and $\boldsymbol{x}_c$ is the corresponding 3D point in canonical FLAME coordinates. $\mathbf{H}_i$ is the head pose, $\mathbf{K}$ and $\mathbf{C}^{w2c}$ are camera intrinsics and extrinsics, and $\mathbf{C}_{ref}$ is the head tracker's reference world origin. Subscripts $w_1$ and $w_2$ refer to the world coordinate systems of capture setups 1 and 2 respectively.

However, what we need to represent in world reference system $w_2$ is not the canonical head $\boldsymbol{x}_c$, but the observed head $\boldsymbol{x}_i$ at frame $i$. 
The relation between canonical pose and real observation is:

\begin{equation}
    \boldsymbol{x}_{c} = \mathbf{H}_{ref}^{-1} \mathbf{H}_i^{-1} \mathbf{C}_{ref} \boldsymbol{x}_{i} .
    \label{eq:e1}
\end{equation}

We write this equation generally as it applies to both world references. 
Note, head poses are applied relative to the first frame ($\mathbf{H}_{ref}=\mathbf{H}_0$). 
We can also write the following relations between the learned head geometry $\boldsymbol{x}_h$, canonical points $\boldsymbol{x}_c$ and observed points $\boldsymbol{x}_i$:

\begin{equation}
    \boldsymbol{x}_h = \mathbf{C}_{ref}^{-1} \mathbf{H}_i^{-1} \mathbf{C}_{ref} \boldsymbol{x}_i ,
\end{equation}
and
\begin{equation}
    \boldsymbol{x}_c = \mathbf{H}_{ref}^{-1} \mathbf{C}_{ref} \boldsymbol{x}_h .
    \label{eq:e2}
\end{equation}

Exploiting the fact that canonical points $\boldsymbol{x}_c$ are are same in both worlds, and referring to Eq.~(\ref{eq:e1}) and~\ref{eq:e2}, we can write:
\begin{equation}
    \mathbf{H}_{ref, w_1}^{-1} \mathbf{C}_{ref, w_1} \boldsymbol{x}_h^{w_1} = \mathbf{H}_{ref,w_2}^{-1}  \mathbf{H}^{-1}_{i,w_2} \mathbf{C}_{ref, w_2} \boldsymbol{x}_{i}^{w_2}.
\end{equation}

This can be rearranged, to obtain the final transformation:
\begin{equation}
    \boldsymbol{x}^{w_2}_{i} = \mathbf{C}_{ref,w_2}^{-1} \mathbf{H}_{i,w_2} \mathbf{H}_{ref,w_2} \mathbf{H}_{ref, w_1}^{-1} \mathbf{C}_{ref, w_1} \boldsymbol{x}^{w_1}_{h}.
    \label{eq:alignment}
\end{equation}

\begin{figure}
\centering
\small
\includegraphics[width=0.45\textwidth]{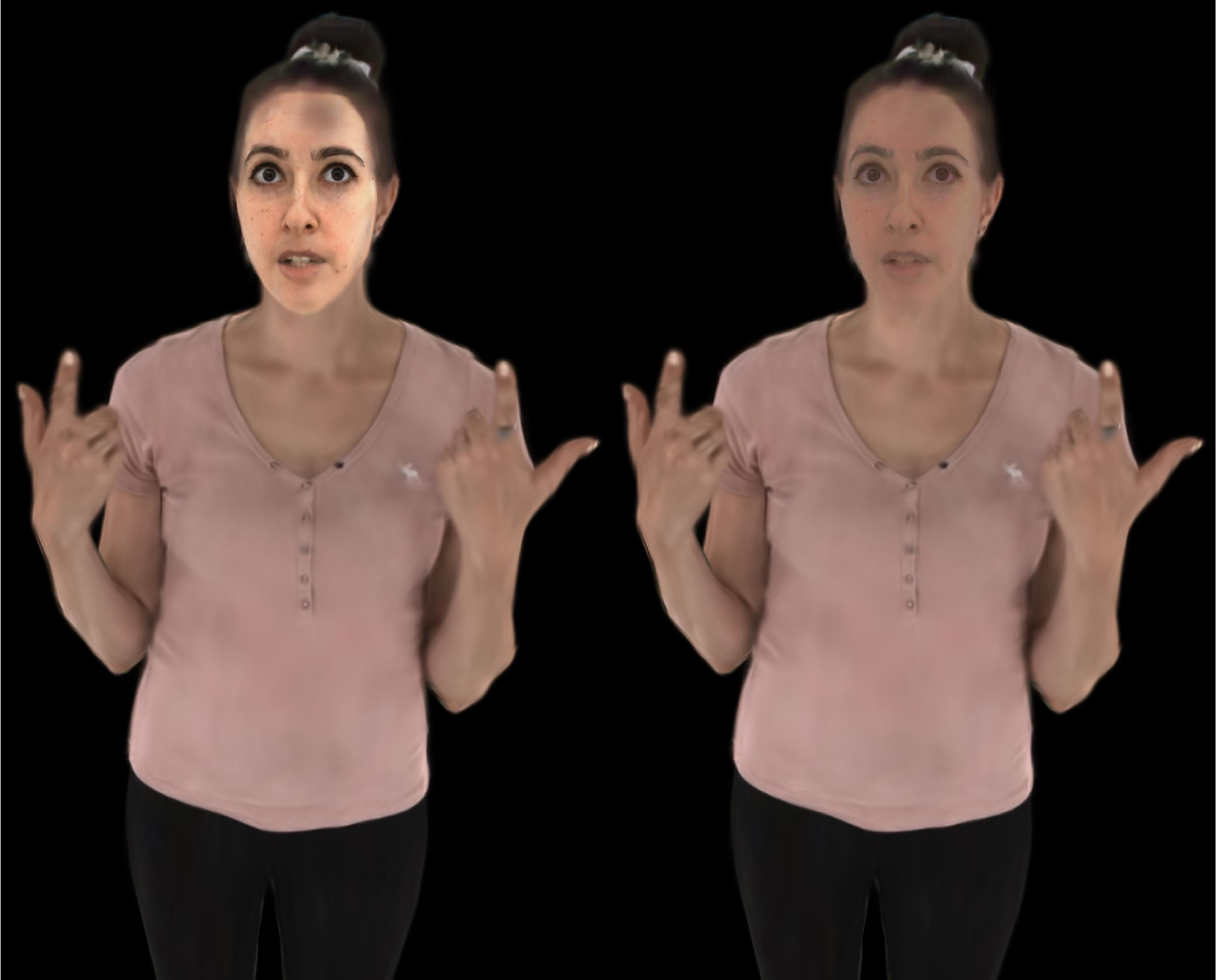}
\caption{\textbf{Cross-setup head-body integration.} The audio-driven head model is trained using HeadGaS++ on the RenderMe-360 dataset~\cite{2023renderme360}, while the body is trained on the same subject from the DNA-Rendering dataset~\cite{cheng2023dna}. Left: the RenderMe-360 head is aligned to the DNA-Rendering body using Eq.~\ref{eq:alignment}. To reduce artifacts, we prune Gaussians from the body model as discussed in Sec.~\ref{sec:head_body_integ}. Right: we optimize for the face colors by un-freezing the last layer of the HeadGaS++ MLP and optimizing jointly with the body model.}
\label{fig:lady_integ}
\end{figure}

Using this transformation we can transform the learned HeadGaS++ geometry $\boldsymbol{x}^{w_1}_{h}$ trained with the head-centric captures from capture 1, into the observed full-body captures $\boldsymbol{x}^{w_2}_{i}$ of capture 2. An example of this alignment is shown in Fig.~\ref{fig:lady_integ}.

\subsection{Language Understanding and Generation}
\label{sec:llm_setup}
To implement our interactive pipeline, we capitalize on the publicly available tools for natural language processing, namely Audio Speech Recognition (ASR), Large Language Models (LLM) and Text-to-Speech (TTS) models. 
We will briefly describe the specific models we used in our framework.

\subsubsection{Audio Speech Recognition (ASR)} 
ASR is the part of the pipeline responsible for understanding input human audio speech and feeding the LLM with the correct question with low latency. We select Whisper~\cite{zhang2018deep}, a state-of-the-art model which demonstrates a strong ability to generalize to many datasets and domains in a zero-shot setting. In our pipeline, we incorporate the Whisper Large V3 model to have support for multiple languages and its distilled version, which is much faster but with support only for English.

\subsubsection{Large Language Model (LLM)} 
One of the most important parts of our pipeline is the LLM module. As latency is a crucial factor for the real-time nature of our application, we need to have a fast and accurate model. 
To this end, we deployed the quantized model of Qwen2 with 0.5B parameters and  Activation-aware Weight Quantization (AWQ)~\cite{lin2024awq}. It is based on the Transformer architecture with SwiGLU activation, attention QKV bias, and group query attention. 

\subsubsection{Text-to-Speech (TTS)}
TTS consists of the last part of our pipeline. For our framework, we deployed the OpenVoice V2~\cite{qin2023openvoice}, which has native multilingual support  (English, Spanish, French, Chinese, Japanese, and Korean).

\section{Implementation Details}
\label{sec:implementation_details}

We implement our method in PyTorch, building upon the 3DGS codebase and its differentiable rasterizer. We initialize the body Gaussian centers using point clouds derived from SMPL-X fittings, while the head points are initialized with 2500 vertices derived from the tracked FLAME mesh~\cite{Zheng2022PointAvatarDP, Zielonka2022InstantVH}. In scenarios where no mesh data is available, the points are sampled randomly within predefined near and far bounds. 

The learning rates for the head MLP, Gaussian positions $\boldsymbol{\mu}$, latent features $\mathbf{F}$, scale $\mathbf{S}$, and rotation $\mathbf{R}$ are set to $1.6 \times 10^{-4}$, $1.6 \times 10^{-4}$, 0.0025, 0.005, and 0.001, respectively, while the body MLP and blending parameters $\boldsymbol{\alpha}$ learning rates are initialized to $1 \times 10^{-4}$. The body is trained with the Adam optimizer, while head optimization is performed using Stochastic Gradient Descent (SGD), with exponential decay scheduling applied to the Gaussian centers $\boldsymbol{\mu}$ and the MLP parameters. The latent feature $\mathbf{F}$ dimensionality is fixed at 32 and initialized to zero. The expression feature vector consists of 32-dimensional audio features from SyncTalk~\cite{peng2024synctalk} and 7-dimensional eye parameters 
from the ARKit tracking~\cite{ARKit}, resulting in 39 elements in total (i.e., $B = 39$). 

The head model is trained on an NVIDIA Tesla V100 GPU for 50,000 iterations, a process that takes roughly one hour to complete. The perceptual loss $L_{p}$ is applied after 10,000 iterations with a weight of 0.1, while the weights for $L_{1}$ and SSIM are set to 0.8 and 0.2, respectively. For the body, each sliding window model is trained for 30,000 iterations in parallel across eight Tesla V100 GPUs. The first 1,000 iterations are designated as a warm-up phase, where only the central frame of each window is trained, with the MLP weights frozen to allow stabilization of the canonical body representation. The final fine-tuning step is performed sequentially on a single GPU for 20,000 iterations per window to ensure temporal consistency. For both models, adaptive densification begins at iteration 500 and concludes by the 15,000th iteration. 

\subsection{Procedural Body Animation}
\label{sec:procedural_animation}
While the dynamic body model produced by SWinGS++ enables replay of the training video sequence, it does not support novel pose generation. To create novel sequences, we use a simple procedural animation technique. At inference time, given speech audio input, our method generates a plausible body-pose sequence to complement the audio-driven facial expressions. The body animation should match the length of the speech. Otherwise, the head and body movements can appear disjointed. 

We do this by defining two states for the avatar: (i) rest pose and (ii) action, each defined by a set of preselected keyframes from the training video. In rest pose, the avatar is in an idle state, waiting to converse with the user when provided input. The rest pose follows an infinite loop, constantly replaying until a user input is detected. The rest pose loop is created by defining start and end keyframes and looping the rendered frames back and forth. Random eye blinks are inserted so that the avatar does not feel lifeless. To create an action sequence, we define a set of keyframes that signify key body movements from the training video, e.g.\,hands gesticulating, carefully selected so that the reversed the motion still feels natural. Our animation method randomly selects from these keyframes to generate a sequence of movements that match the length of the input audio, finishing at the rest pose end keyframe so that the movement smoothly transitions back into the idle state.

\subsection{Interactive Viewer} 
\label{sec:interactive_viewer}

In this section, we detail our interactive viewer implementation, serving as the central platform for integrating our prebuilt avatar models with interactive Text-to-Speech (TTS) and Large Language Model (LLM) functionalities. The system combines these components to deliver a smooth and immersive conversational human avatar experience. We build upon the original viewer from 3DGS~\cite{kerbl3Dgaussians}, which excels in rendering high-fidelity static scenes but lacks dynamic and interactive capabilities. To bridge this gap, we extend it to support real-time interactivity and conversation by adding modules for: (1)~text or microphone input, (2) Automatic Speech Recognition (ASR), (3) LLM processing, (4) TTS audio generation, (5) audio-to-SyncTalk expression parameter generation, and (6) our HeadGaS++ and dynamic body models.

The key features of our interactive viewer include:

\subsubsection{Modular Integration with a Remote Server}
The viewer adopts a modular architecture allowing for integration with remote server modules, including the ASR, LLM, TTS, and SyncTalk components. Each component functions independently while maintaining synchronized communication between viewer and server.
By keeping the ASR, LLM, TTS, and SyncTalk modules on a remote server, we minimize the computational load on the local viewer. The local viewer focuses solely on rendering and minimal communication, including:
\begin{itemize}
    \item Sending: user inputs (audio from a microphone or text).
    \item Receiving: processed outputs (ASR text, LLM text responses, TTS audio, SyncTalk parameters).
    \item HeadGaS++ inference: using SyncTalk and head pose inputs to predict Gaussian attributes (Sec.~\ref{sec:head_model}).
    \item Rendering: background, body, and face Gaussians. 
\end{itemize}
This setup minimizes communication between the local viewer and the remote server to essential data exchanges, while computationally intensive processes run concurrently on the remote server. The final output from the server is the SyncTalk expression parameters, which the local viewer feeds into the embedded HeadGaS++ module (using LibTorch). Frame-wise Gaussian parameters are generated to produce audio-synchronized facial expressions.

\subsubsection{Real-Time Synchronization}
The viewer synchronizes facial expressions, lip movements, and body gestures with generated speech in real time. Given a user input, the TTS module converts a LLM-generated text response into an audio signal, driving the facial animation. Our procedural animation module (Sec.~\ref{sec:procedural_animation}) generates natural idle motion to ensure the avatar displays natural body movement (e.g. eye-blinking) while awaiting user input or server response. 
To smoothly transition between idle and talking states, audio playback and facial animation start simultaneously. The avatar's Gaussians are rendered at fixed 30 FPS to synchronize with the audio, while background Gaussians utilize the remaining processing capacity, enabling our system to run at \textbf{105.27 FPS}.

\section{Experimental Results}\label{sec:experiments}
This section presents visual and quantitative results supporting our method. We recommend that readers inspect the supplementary video, which shows high-quality renderings and several user interaction scenarios with our avatar. The model presented in this video was captured using a synchronized multi-view system with 24 cameras, recording a subject performing speech and gestures simultaneously. To our knowledge, no public dataset exists containing similar data (both speech and body motion). As a result, we evaluate our body and face models separately using dedicated datasets in Sec.~\ref{sec:body_eval} and \ref{sec:head_eval}. As these datasets have been released by the same group, we select sequences that depict the same subject in both datasets, enabling us to build an interactive avatar from different captures. This procedure is described in Sec.~\ref{sec:cross_setup}.

\subsection{Body Model Evaluation}
\label{sec:body_eval}

\begin{figure}
\centering
\small
\subfloat[DNA-Rendering]{\includegraphics[width = 3.0in]{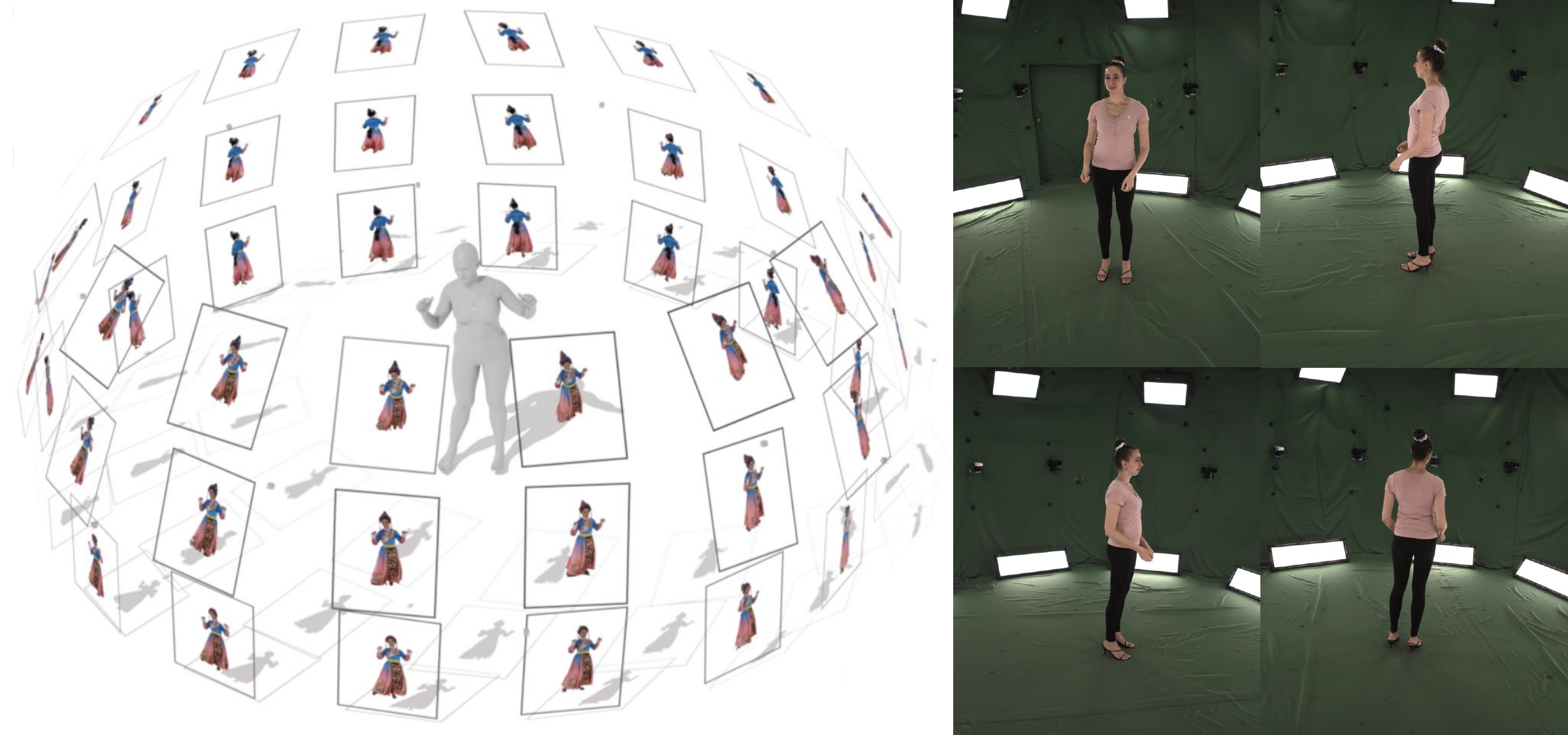}}
\hfill
\subfloat[RenderMe-360]{\includegraphics[width = 2.5in]{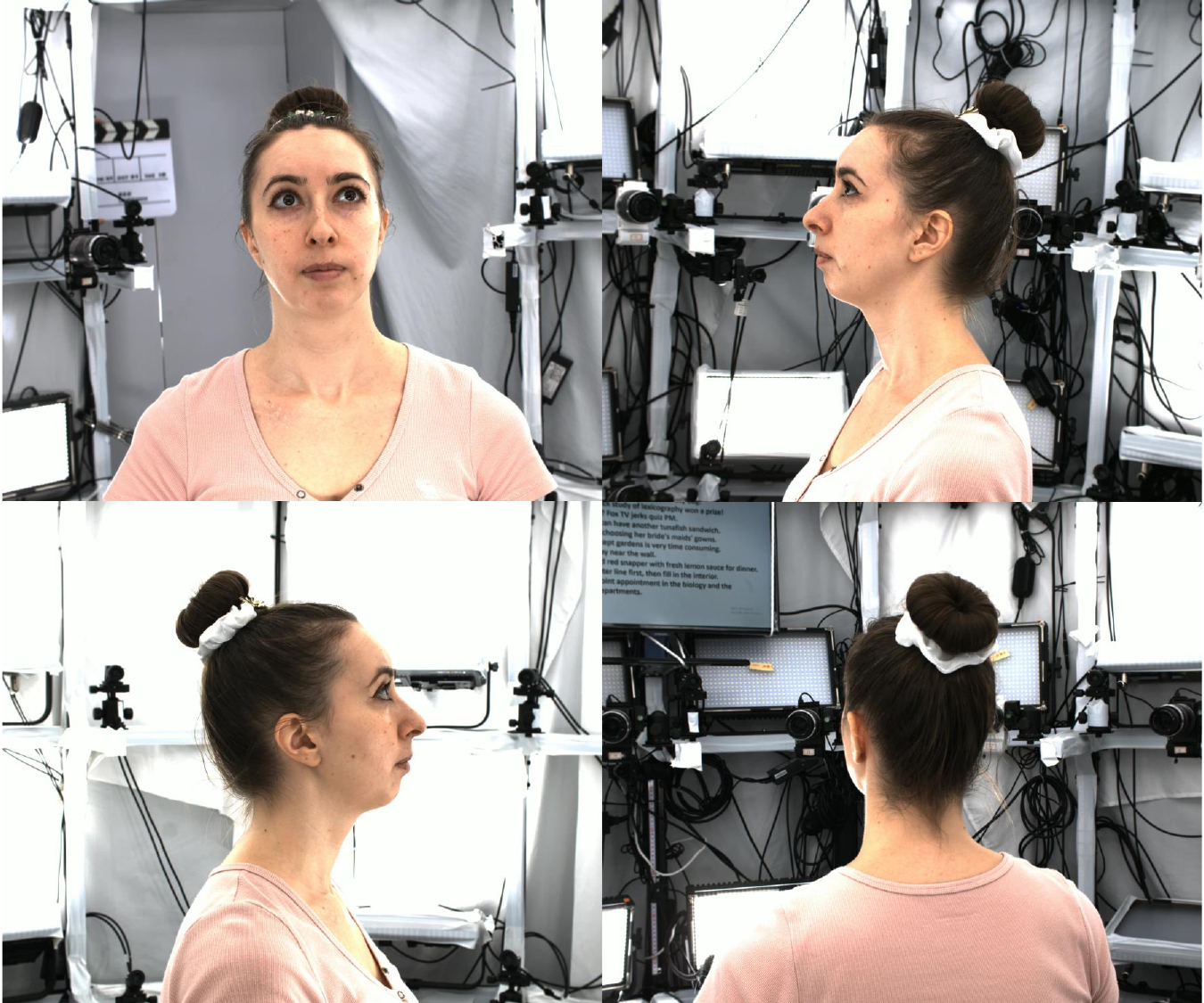}}
\caption{\textbf{Datasets used for cross-setup avatar creation.} We can use data of the same subject captured from two different environments to train our Gaussian avatar. We used DNA-Rendering for the body model and RenderMe-360 for the head model. The alignment procedure is discussed in section~\ref{sec:cross_setup}.}
\label{fig:datasets}
\end{figure}

We use the public DNA-Rendering dataset~\cite{cheng2023dna} to evaluate our proposed body model. This dataset contains full body human performances recorded by 60 synchronized cameras at 15 FPS in a light stage environment. Despite dense camera coverage, this is a challenging dataset due to complex human motion, loose clothing with rich textures, a large diversity of scanned individuals (ethnicity, age, height, gender, etc.), and various object interactions. An overview of the camera positions and images is given in Fig.~\ref{fig:datasets}.


\subsubsection{Novel View Synthesis Evaluation}
Following prior work, we evaluate the quality of our model through novel view synthesis (NVS). DNA-Rendering~\cite{cheng2023dna} includes an NVS benchmark but does not provide the exact data splits to enable direct comparison to their results. However, we replicate it as closely as possible while comparing to the most recent baselines.

We train one model for each sequence of 225 time steps using 48 cameras evenly distributed around the subject. 12 cameras are held out for evaluation. We use the six sequences shown in \href{https://dna-rendering.github.io/inner-benchmark-novel-view.html}{the benchmark webpage}. To save computation, we preprocess training images by cropping based on segmentation masks to avoid rendering background pixels. Evaluation is done using PSNR, SSIM~\cite{wang2004image}, and LPIPS~\cite{zhang2018unreasonable} metrics. PSNR-M refers to masked PSNR, i.e. PSNR computed only on the ground truth human segmentation mask. We compare our body model SWinGS++ (Sec.~\ref{sec:body_model}) to the following baselines:
\begin{itemize}
    \item SWinGS~\cite{shaw2024swings}: our prior method without spatial-temporal feature encoders.
    \item 4D Gaussian Splatting~\cite{Wu_2024_CVPR}: a 4D model that proposed the spatial-temporal feature encoder for 4DGS. In contrast with our method, it does not use sliding windows.
    \item SpaceTime Gaussians~\cite{Li_STG_2024_CVPR}: a 4D model using time-conditioned parametric functions to model the evolution of Gaussians.
    \item HuGS~\cite{moreau2024human}: An animatable GS model, i.e. deformations are driven by the SMPL-X~\cite{SMPL-X:2019} skeleton (Sec.~\ref{sec:hugs_comparison}).
\end{itemize}

\begin{figure*}
\centering
\small
\includegraphics[width=0.85\textwidth]{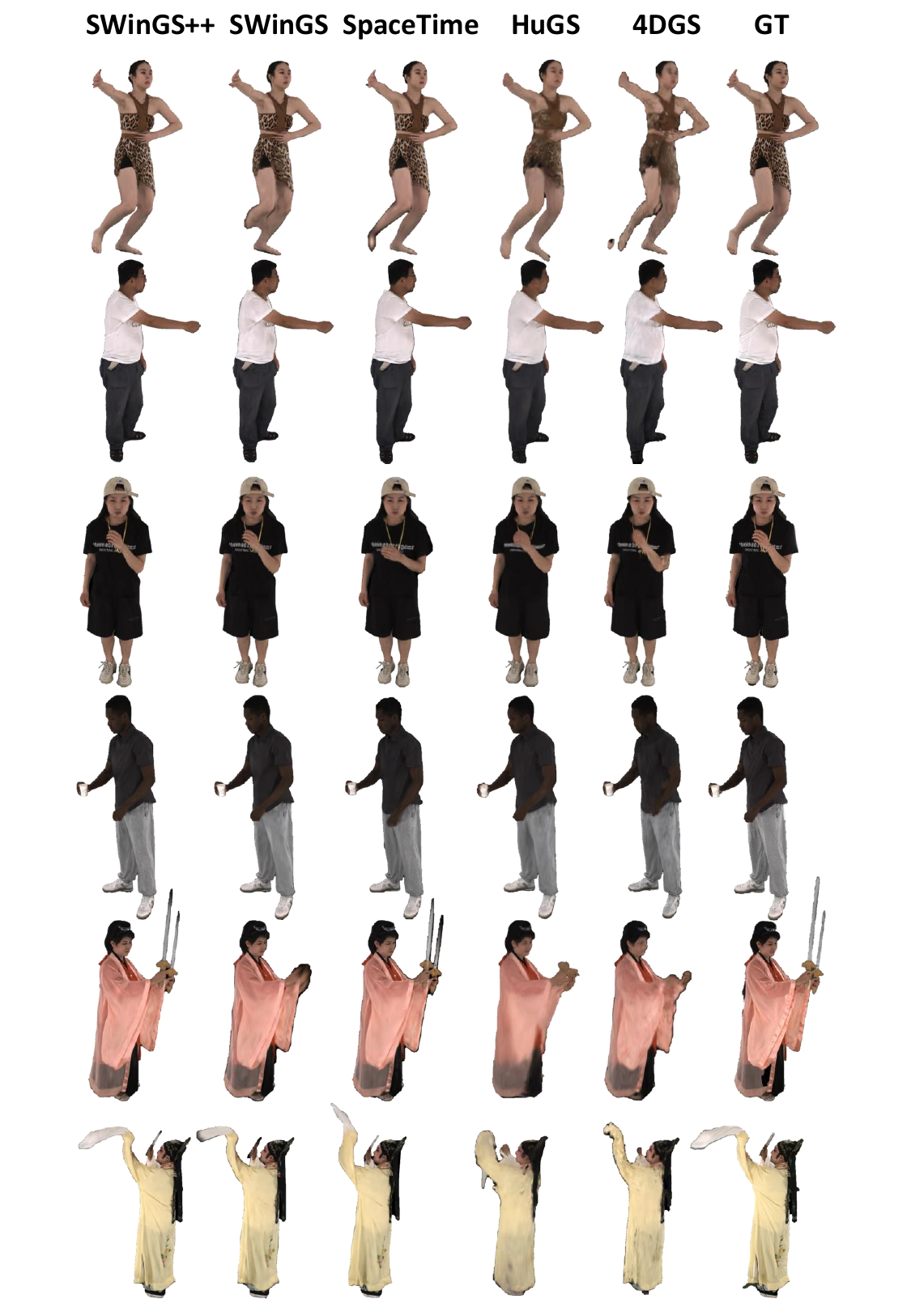}
\caption{\textbf{Qualitative comparison of novel view synthesis.} Our method SWinGS++ displays notably sharper rendering with better motion reconstruction.}
\label{fig:novel_view}
\end{figure*}

Quantitative comparison with all other methods is shown in Table~\ref{tab:body_nvs} and qualitative comparisons are provided in Fig.~\ref{fig:novel_view}. We observe that fitting this dataset is difficult for most methods. Spacetime Gaussians~\cite{Li_STG_2024_CVPR} is  limited to short sequences as it is memory demanding; we had to split the video into 5 segments of 45 frames and train a dedicated model for each. 4DGaussians~\cite{Wu_2024_CVPR} is compatible with longer sequences but presents quite low rendering quality with poor detail reconstruction. HuGS~\cite{moreau2024human} supports training on long sequences but, as it relies on a template-body model, it is not able to capture objects and loose clothing. It is generally more blurry overall, due to noise in the input body pose parameters. SWinGS~\cite{shaw2024swings} and SWinGS++ perform better than these baselines thanks to the sliding windows approach. Finally, the spatial-temporal feature encoder of SWinGS++ improves the rendering quality further by 0.7dB PSNR and better reconstructs faster movements, e.g. the subject's leg in Fig.~\ref{fig:novel_view} row 1.

\begin{table}
\centering
\setlength{\tabcolsep}{2pt}
\caption{Quantitative results of the body reconstruction for novel view synthesis on the DNA Rendering dataset~\cite{cheng2023dna}. We report PSNR, masked PSNR (PSNR-M), SSIM and  LPIPS metrics.}
\label{tab:body_nvs}
\begin{tabular}{ l c c c c } 
\hline
Method & PSNR $\uparrow$ & PSNR-M $\uparrow$ & SSIM $\uparrow$  & LPIPS $\downarrow$ \\
\hline

\textbf{SWinGS++ (ours)} & \textbf{30.1701} & \textbf{23.6950} & \textbf{0.9699} & \textbf{0.0807} \\ 
SWinGS~\cite{shaw2024swings} & 29.4488 & 22.9739 & 0.9647 & 0.0902 \\ 
SpacetimeGauss.\cite{Li_STG_2024_CVPR} & 29.3039 & 22.8494 & 0.9621 & 0.0924 \\
4DGS~\cite{Wu_2024_CVPR} & 26.8197 & 20.2145 & 0.9441 & 0.1152 \\
HuGS~\cite{moreau2024human} & 25.0400 & 18.9540 & 0.9404 & 0.1118 \\
\hline
\end{tabular}
\end{table}

\subsubsection{Comparison with Animatable Body Models}
\label{sec:hugs_comparison}

\begin{figure}
\centering
\small
\subfloat[SWinGS++]{\includegraphics[width = 1.0in, decodearray={0.08 1 0.08 1 0.08 1}]{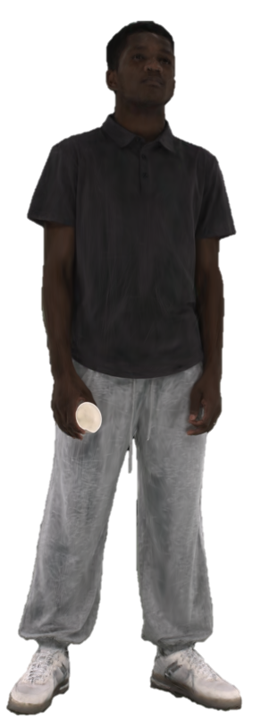}}
\subfloat[HuGS]{\includegraphics[width = 1.0in, decodearray={0.08 1 0.08 1 0.08 1}]{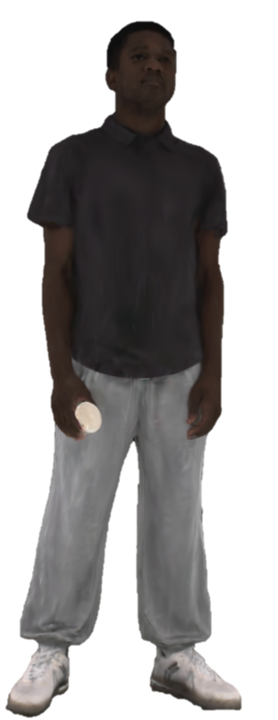}}
\subfloat[Novel pose]{\includegraphics[width = 1.0in, decodearray={0.08 1 0.08 1 0.08 1}]{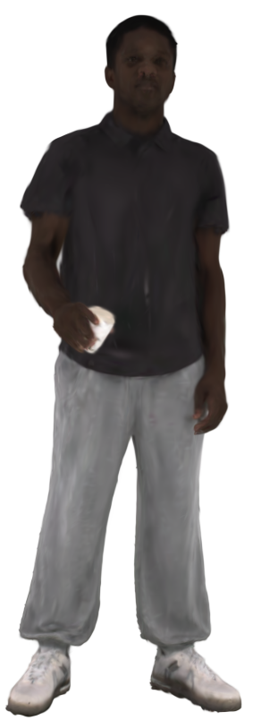}}
\caption{\textbf{Comparison between our dynamic body model SWinGS++ and HuGS.} SWinGS++ provides notably better novel view synthesis results than HuGS on observed motion. However, HuGS also enables novel pose synthesis (c) which is not possible with a 4D reconstruction model like SWinGS++.}
\label{fig:novel_pose}
\end{figure}

As presented in Sec.~\ref{sec:body_model}, SWinGS++ learns a time-dependent Gaussian Splatting model for the dynamic human. However, another approach is to use an animatable model where 3D Gaussians are skinned to a human skeleton~\cite{pang2024ash,li2024animatable,hu2024gaussianavatar,wen2024gomavatar,jung2023deformable}. 
The advantage is the ability to render the subject in any target pose, rather than being limited to replay of the training motions. We explored this using HuGS~\cite{moreau2024human} for the body of our avatar. However, we observed that HuGS has reduced render quality compared to SWinGS++, although it is able to extrapolate to novel gestures with reasonable quality, as shown in Fig.~\ref{fig:novel_pose}. 
While the render quality is reduced compared to 4D methods, we note that this benchmark contains human-object interactions (e.g. holding a sword) which cannot be represented by a human body model only. 

Given the difference in render quality, we decided to build our interactive avatar using SWinGS++ to maximize photorealism and generate a convincing avatar. However, 4D models limit us to the procedural body animation described in Sec.~\ref{sec:procedural_animation} and thus the motion of the avatar can sometimes look repetitive or not well adapted to the speech. Animatable Gaussian-based avatars would enable driving the body based on speech input using 3D motion synthesis methods~\cite{habibie2022motion, mir2024generating}. We see this direction as a promising future work.

\subsection{Head Model Evaluation}
\label{sec:head_eval}

To evaluate our proposed animatable head model, we follow~\cite{li2024talkinggaussian}, using four high-definition talking portrait video clips from previous
publicly released video sets (three males, one female) for testing. Each portrait video consists of 6,500 frames on average that are divided into training and test sets in a ratio of $10:1$. The video clips are cropped and resized to $512 \times 512$, apart from the Obama video, which has a resolution of $450 \times 450$. 

\begin{figure*}
\centering
\small
\includegraphics[width=1\textwidth]{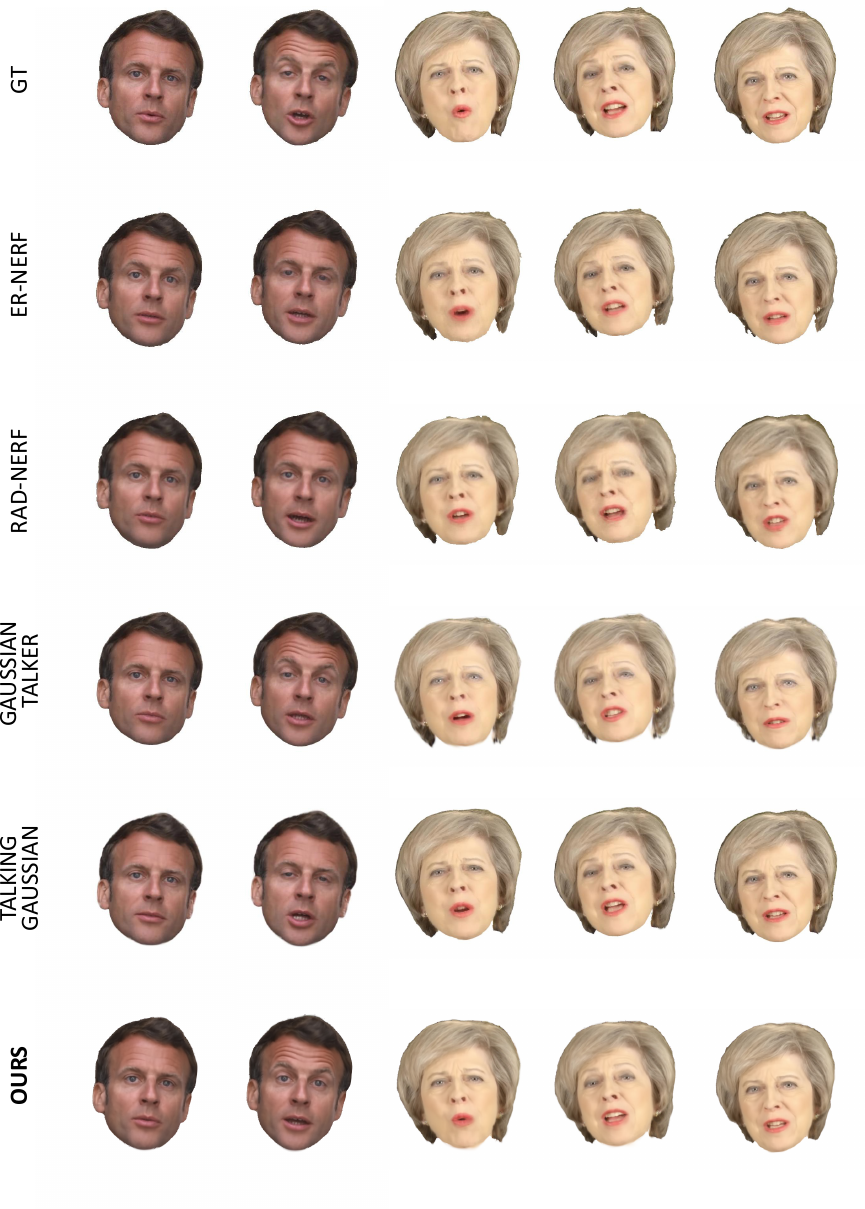}
\caption{\textbf{Qualitative comparison of lip synchronization.} Our method HeadGaS++ displays better performance in synthesizing synchronized talking heads compared to the baseline methods. Please zoom in for better visualization.}
\label{fig:head_qualitative}
\end{figure*}

We compare our model with two state-of-the-art talking head NeRF-based methods, namely ER-NeRF~\cite{li2023ernerf}, RAD-NeRF~\cite{tang2022radnerf}, and two 3D Gaussian Splatting-based methods, TalkingGaussian~\cite{li2024talkinggaussian} and GaussianTalker~\cite{cho2024gaussiantalker}. The evaluation settings for the reconstruction quality and lip-audio synchronization ability are as follows:
\begin{enumerate}
\item \textit{The self-reconstruction setting:} Each video is split into training and test sets. We trained each model using the audio, expression parameters, and head-pose sequences from the training set. For testing, the model synthesizes new instances of talking face using input audio, expressions, and pose sequences from the unseen test set. PSNR, SSIM~\cite{wang2004image}, LPIPS~\cite{zhang2018unreasonable}, LMD~\cite{chen2018lip} are used in quantitative comparison. Lip synchronization is evaluated using the confidence score (Sync-C) and error distance (Sync-E) from SyncNet~\cite{son2017lip, chung2017out}. 

Table~\ref{tab:head_quantitative}  shows the quantitative comparison with all other methods and qualitative comparisons are provided in Fig.~\ref{fig:head_qualitative}. Our method outperforms the state-of-the art methods in rendering quality, motion quality, and efficiency. It achieves the highest PSNR and SSIM, indicating superior image fidelity and structural similarity, while maintaining a competitive LPIPS score. In motion quality, it leads with the highest Sync-C, reflecting better synchronization. Efficiency-wise, it achieves the highest FPS (250) with a reasonable training time of 1 hour, making it highly suitable for real-time applications. The quality of the reconstructed faces can be seen in the qualitative results in Fig.~\ref{fig:head_qualitative}.

\item \textit{The cross-driven setting:}
In this case, each pre-trained model is driven by audio tracks from other videos. Specifically, we used the same audio samples as in the previous work SynObama~\cite{suwajanakorn2017synthesizing} to evaluate the ``Macron'' and ``Lieu'' portraits.  Table~\ref{tab:head_cross} highlights the effectiveness of our method. It achieves high scores in Sync-C even in cross-lingual scenarios. Despite the Macron model being trained in French, it scores the lowest Sync-C by a considerable margin, indicating superior alignment of our lip movements. This emphasizes our proposed method’s robustness and adaptability in handling cross-lingual lip synchronization, outperforming the other methods.
\end{enumerate}

\begin{table*}[t]
\centering
\caption{Quantitative results on head reconstruction quality and lip-synchronization for the \textit{self-reconstruction setting}. We compare to different methods and report rendering quality, motion quality, and efficiency. Each model is trained using the audio, expression parameters, and head-pose sequences from the training set and tested on unseen input audio, expressions, and poses.}
\label{tab:head_quantitative}
\resizebox{\textwidth}{!}{
\begin{tabular}{lccccccc}
\toprule
\textbf{Methods} & \multicolumn{3}{c}{\textbf{Rendering Quality}} & \multicolumn{2}{c}{\textbf{Motion Quality}} & \multicolumn{2}{c}{\textbf{Efficiency}} \\
\cmidrule(lr){2-4} \cmidrule(lr){5-6} \cmidrule(lr){7-8}
& PSNR $\uparrow$ & SSIM $\uparrow$ & LPIPS $\downarrow$  & LMD $\downarrow$ & Sync-C $\uparrow$ & Training Time & FPS \\
\textbf{Ground Truth} & N/A & 0 & 1.000 & 6.861 &  7.584 & - & - \\
\hline
RAD-NeRF~\cite{tang2022radnerf} & 26.794 & 0.901 & 0.083 & 2.907 & 4.988 & 3h & 25 \\
ER-NeRF~\cite{li2023ernerf} & 27.350 & 0.904 & 0.063 & 2.836 & 5.172 & 1h & 35\\
TalkingGaussian~\cite{li2024talkinggaussian} & 29.317 & 0.920 & \textbf{0.046} & \textbf{2.688} & 5.802 & 0.8h & 110 \\
GaussianTalker~\cite{cho2024gaussiantalker} & 29.127 & 0.911 & 0.085 & 2.814 & 5.350 & 5h & 130\\
\hline
\textbf{Ours} & \textbf{30.398} & \textbf{0.935} & 0.051 & 2.793 & \textbf{5.928} & 1h & 250\\
\bottomrule
\end{tabular}
}
\end{table*}

\begin{table*}[t]
\centering
\caption{Quantitative results on lip synchronization for the \textit{cross-driven setting}. Following previous works, we extract two audio clips from the SynObama demo~\cite{suwajanakorn2017synthesizing} to drive each method and evaluate lip synchronization using the confidence score (Sync-C) and error distance (Sync-E) from SyncNet~\cite{son2017lip, chung2017out}.}
\label{tab:head_cross}
\resizebox{\textwidth}{!}{
\begin{tabular}{lcccccccccccc}
\toprule
\textbf{Methods} & \multicolumn{4}{c}{Test Audio A} & \multicolumn{4}{c}{Test Audio B} \\
\cmidrule(lr){2-5} \cmidrule(lr){6-9}
& \multicolumn{2}{c}{``Macron''} & \multicolumn{2}{c}{``Lieu''} & \multicolumn{2}{c}{``Macron''} & \multicolumn{2}{c}{``Lieu''} \\
\cmidrule(lr){2-3} \cmidrule(lr){4-5} \cmidrule(lr){6-7} \cmidrule(lr){8-9}
& Sync-E $\downarrow$ & Sync-C $\uparrow$ & Sync-E $\downarrow$ & Sync-C $\uparrow$ & Sync-E $\downarrow$ & Sync-C $\uparrow$ & Sync-E $\downarrow$ & Sync-C $\uparrow$ \\
\textbf{Ground Truth} & 0 &  7.463 & 0 &  7.463  & 0 & 7.37 & 0 & 7.37 \\
\hline
RAD-NeRF~\cite{tang2022radnerf} & 7.999 & \textbf{6.419} & 9.910 &4.051 & 7.875 &6.894 & 8.728 &6.135 & \\
ER-NeRF~\cite{li2023ernerf} & 8.618 &6.110 & 11.105 &2.828 & 7.826 & \textbf{7.271} & 11.241 &3.168 & \\
TalkingGaussian~\cite{li2024talkinggaussian} & 8.597 &6.038 & 10.389 &3.242 & 8.008 &6.803 & 10.429 &3.736 & \\
GaussianTalker~\cite{cho2024gaussiantalker}  & 9.381 &5.399 & 10.691 &3.138 & 8.738 &6.522 & 11.142 &3.285 & \\
\textbf{Ours} & \textbf{7.566} & 6.235 & \textbf{7.431} & \textbf{6.487} & \textbf{7.231} & 7.251 & \textbf{7.463} & \textbf{7.491} & \\
\bottomrule
\end{tabular}
}
\end{table*}

\subsection{Computational Speed and Complexity}
\label{subsec: computation and complexity}
As we aim to achieve a real-time interactive conversational avatar, we evaluate our viewer’s performance on two different systems: a Linux and a Windows system with different hardware configurations, to assess computational speed and efficiency. On the Linux system running Ubuntu equipped with an Intel Core i9-9900X CPU, NVIDIA GeForce RTX 2080 GPU, and 64 GB of RAM, we achieve an average frame rate of 70.60 FPS when rendering the head, body, and background (BG) together—equivalent to 14.16 ms per frame. On the Windows system running Windows 11, equipped with an Intel Core i9-13900HX CPU, NVIDIA GeForce RTX 4090 Laptop GPU, and 32 GB of RAM, we achieved an average frame rate of~\textbf{105.27 FPS} under the same conditions—equivalent to~\textbf{9.50 ms} per frame. These results confirm that our system operates well above the typical threshold for real-time performance (30 FPS), ensuring smooth and responsive interactions. Detailed performance metrics are presented in Table~\ref{tab:viewer_performance}.

\begin{table}[h]
\centering
\footnotesize
\setlength{\tabcolsep}{4pt}
\caption{Performance metrics of our interactive viewer on two system configurations. A \cmark\ indicates inclusion of a component; absence of a mark indicates exclusion.}
\label{tab:viewer_performance}
\begin{tabular}{ c c c c c }
\toprule
\textbf{BG} & \textbf{Body} & \textbf{Head} & \multicolumn{2}{c}{\textbf{FPS (ms)}} \\
& & (HeadGaS++) & \textbf{Linux} & \textbf{Windows} \\
\midrule
\cmark &       &       & 125.68 (7.96)  & 169.92 (5.89)  \\
\cmark & \cmark &       & 114.86 (8.70)  & 152.23 (6.57)  \\
\cmark &       & \cmark & 74.60  (13.40) & 112.53 (8.89)  \\
\cmark & \cmark & \cmark & 70.60  (14.16) & 105.27 (9.50)  \\
\bottomrule
\end{tabular}
\end{table}

\section{Conclusion}
\label{sec:conclusions}

We have presented \textit{Interactive Conversational 3D Virtual Human} (ICo3D), a comprehensive system for creating interactive, photorealistic 3D virtual human avatars. Our system seamlessly integrates dynamic full-body 3D reconstruction, dynamic facial expressions, and audio-driven conversational capabilities. We address limitations in the state-of-the-art by unifying novel methods for audio-driven facial animation, LLM-powered conversational interaction, and dynamic human rendering. Additionally, our method enables real-time rendering of avatars with free-viewpoint camera navigation, delivering an immersive and interactive user experience.

Our experimental results demonstrate the effectiveness of ICo3D in producing lifelike avatars with high-quality 3D reconstructions surpassing state-of-the-art methods. Our extended body reconstruction method SWinGS++ improves upon its predecessor with better motion estimation and sharper reconstructions, even with significant human motion. Similarly, our extended audio-driven 3D head model HeadGaS++ outperforms state-of-the-art methods in terms of rendering quality and lip synchronization in both self-driven and cross-driven settings. Qualitative results show our avatars are capable of engaging in natural, LLM-driven conversations, with precise synchronization between audio and realistic facial expressions. 


\backmatter

\bmhead{Supplementary information}

Please refer to our supplementary video for more visualizations and results.

\section*{Declarations}

\subsection*{Data Availability}
The findings and experimental section of this paper are based on multiple publicly accessible datasets which are freely available. These include the DNA-Rendering dataset~\cite{cheng2023dna}, which can be accessed at \url{https://dna-rendering.github.io/}, the RenderMe-360 dataset~\cite{2023renderme360}, available at \url{https://renderme-360.github.io/}, and the TalkingGaussian dataset~\cite{li2024talkinggaussian}, available at \url{https://fictionarry.github.io/TalkingGaussian/}. Requests for access can be made to the corresponding authors, subject to approval in accordance with applicable regulations. Additionally, this manuscript includes visualizations on data not publicly available due to internal privacy regulations.

\bibliography{sn-bibliography}

\end{document}